\documentclass[runningheads]{llncs}
\usepackage{booktabs,tabularx,graphicx}
\newcolumntype{C}{>{\centering\arraybackslash}X}

\usepackage{tikz}
\usepackage{comment}
\usepackage{amsmath,amssymb} 
\usepackage{color}
\usepackage{multirow}

\definecolor{mygreen}{RGB}{0, 120, 0}

\usepackage{amsmath}
\newcommand*\textfrac[2]{
  \frac{\text{#1}}{\text{#2}}
}
\begin{document}
\pagestyle{headings}
\mainmatter
\def\ECCVSubNumber{11}  

\title{An Efficient Method for Face Quality Assessment on the Edge
       } 

\titlerunning{An Efficient Method for Face Quality Assessment on the Edge}
%
\author{Sefa Burak Okcu\orcidID{0000-0003-0308-7394}  \and
Burak Oğuz Özkalaycı\orcidID{0000-0002-7690-2751} \and
Cevahir Çığla\orcidID{0000-0003-3246-3176} }
\authorrunning{S. B. Okcu et al.}
%
\institute{Aselsan Inc., Turkey \\
\email{\{burakokcu,bozkalayci,ccigla\}@aselsan.com.tr}}
\maketitle

\begin{abstract}
Face recognition applications in practice are composed of two main steps; face detection and feature extraction. In a sole vision-based solution, the first step generates multiple detections for a single identity by ingesting a camera stream. A practical approach on edge devices should prioritize these detections of identities according to their conformity to recognition. In this perspective, we propose a face quality score regression by just appending a single layer to a face landmark detection network. With almost no additional cost, face quality scores are obtained by training this single layer to regress recognition scores with surveillance like augmentations. We implemented the proposed approach on edge GPUs with all face detection pipeline steps, including detection, tracking, and alignment. Comprehensive experiments show the proposed approach’s efficiency through comparison with state-of-the-art face quality regression models on different data sets and real-life scenarios. 

\end{abstract}

\section{Introduction}
 
In the COVID-19 pandemic, the need and importance of remote capabilities during access control and thermography for human screening have increased. Face detection and recognition systems are the most popular and endeavored tools for remote and isolated capabilities in such scenarios. Moreover, hardware improvements and publicly available face data sets lead to increased demand for face recognition (FR) research. The FR systems can be categorized into two mainstreams according to the application area: access control and payment involving volunteered faces \cite{KonenZnFace},\cite{Yugashini2013},\cite{Ibrahim2011},\cite{Feng2017} and surveillance for public safety \cite{Wheeler2010},\cite{Lei2009},\cite{Xu2014},\cite{Haghighat2017}. People try to provide recognizable faces in the systems for the first scenario where the environment is controlled in terms of lighting, head pose and motion. FR problem is mostly solved for this scenario such that access control devices are available in many facilities, and millions of people use smartphones with face recognition technology. On the other hand, new problems form for these scenarios that need special attention such as liveness detection to handle spoofing attacks \cite{Chingovska2016},\cite{Erdogmus2014},\cite{Akbulut2017} or masked faced recognition under severe conditions. 

FR for surveillance involves a variety of open problems compared to the first group. In this type, cameras observe people that walk-run through streets, terminals in their natural poses indoors and outdoors. People occlude each other, lighting is not controlled and people do not look at the cameras intentionally. Face detection, mostly solved for specifically large and clear faces, encounters these uncontrolled problems as the first step. The cameras should be located appropriately to catch the faces with sufficient resolution (e.g. 60 pixels between two eyes) and as frontal as possible. During the face detection step, many unrecognizable face photos of an individual can be captured that requires additional filtering to boost FR rates. Apart from that, face databases are in the order of millions for public security that makes FR a harder problem.

Many researchers put their attention on developing new cost functions and convolutional neural networks for high performance FR algorithms \cite{SphereFace2017},\cite{CosFace2018},\cite{ArcFace2018}, \cite{MobileFacenet2018},\cite{Vargfacenet2019}. At the same time, lightweight face detection gains attention of researchers \cite{Mtcnn2016},\cite{SSH2017},\cite{S3fd2017},\cite{RetinaFace2019} that enables algorithms to run on edge devices with high performance in real-time. Following face detection on the edge, FR is generally performed on servers to cope with large databases and the need for handling complex scenes. This solution requires the transfer of detected faces from edge devices to servers. At that point, face quality assessment (FQA) methods \cite{Abaza},\cite{Vignesh-2015},\cite{QiCNN},\cite{Lienhard-2015},\cite{HernandezOrtega2019} is the key for selecting the best faces to reduce the amount of data transfer and decrease the false alarm rate in FR. In this approach, multiple faces of an individual, obtained during face detection and tracking, are analyzed and the most appropriate faces are selected for the FR step. FQA is an important feature that is applicable on both the edge and server side and removes redundant, irrelevant data in an FR system. 

In this study, we focus on efficient FQA that is achieved in conjunction with face landmark detection on the edge devices. Giving the related work for FR and FQA in the next section, we provide the motivation and contribution of this study that is followed by the details of the proposed face quality measure framework in the fourth section. After the comprehensive experiments on the reliability of the proposed approach and correlation with the FR rates, we conclude in the final section.
    
\section{Related Work}

In the last decade, various approaches have been proposed to address FQA which can mainly be categorized into classical and machine learning techniques. The classical techniques rely on several image statistics while machine learning methods exploit convolutional neural networks to train FQA models. \cite{Wong}  defines an ‘ideal’ face using patch-based local analysis by taking alignment error, image shadowing and pose variation into consideration simultaneously. \cite{Abaza} combines the classical image quality metrics of contrast, brightness, focus, An Efficient Method for Face Quality Assessment on the Edge 3 sharpness and illumination for FQA. \cite{Lienhard-2014} propounds that foreground-background segmentation improves image quality assessment performance, especially for frontal facial portraits. In \cite{Lienhard-2015}, support vector machine is utilized to model the quality of frontal faces based on 15 features extracted for different image regions (entire image, face region, eyes region, mouth region) according to sharpness, illumination, contrast and color information. In \cite{Chen2015FaceIQ}, a learning to rank based framework is proposed to evaluate face image quality. Firstly, weights are trained to calculate scores from five different face feature vectors. These scores are combined using a second-order polynomial in order to obtain the ultimate face quality score. \cite{Vignesh-2015} expresses a convolutional neural network (CNN) based FQA algorithm which is trained to imitate face recognition ability of local binary patterns or histogram of oriented gradients based face recognition algorithms and predict a face quality score. \cite{Anantharajah} utilizes a technique for high quality frame selection along with video by fusing different normalized face feature scores which are face symmetry, sharpness, contrast, brightness, expression neutrality and openness of the eyes via a two layers neural network. 

In \cite{Li2016ImgQ}, a deep feature-based face quality assessment is proposed. After deep-features are extracted in face images, the best representative features are selected using sparse coding and face quality score obtained from support vector regression (SVR). \cite{Khastavaneh2017FaceIQ}  a deep feature-based face quality assessment is proposed. After deep features are extracted in face images, the best representative features are selected using sparse coding and face quality score obtained from support vector regression (SVR). \cite{QiCNN} proposed a CNN based key frame engine which is trained according to recognition scores of a deep FR system. In a similar fashion, \cite{QiBoost} proposed a CNN based FQA module trained according to recognition scores obtained by cosine similarity of extracted deep face features in a database. Nevertheless, it is not obvious whether the best scores are attained with the same identities or not. \cite{Best-Rowden}compares two face image quality assessment methods which are designed by SVR model on features extracted by a CNN with two different labelling methods: human quality values (HQV) assigned by humans via Amazon Turk and matcher quality values obtained by similarity scores between faces. The models are tested on two unconstrained face image databases, LFW and IJB-A, and it is concluded that HQV is a better method for quality assessment. In \cite{YuFace}, Yu et al. designed a shallow CNN in order to classify images into degradation classes as low-resolution, blurring, additive Gaussian white noise, salt and pepper noise and Poisson noise. The model was trained on the augmented CASIA database. They calculated a biometric quality assessment score by weighted summation of degradation class probabilities with the weights of each class’s recognition accuracy. \cite{Dhar} indicates an approach in order to obtain iconicity score based on a Siamese multi-layer perceptron network which is trained via features extracted from face recognition networks. 

In \cite{HernandezOrtega2019}, a ResNet-based CNN model, called FaceQnet, is proposed. The quality values are obtained by the similarity scores between gallery faces selected by ICAO compliance software and other faces in the database. They observed that face quality measurement extracted from FaceQnet improved performance of face recognition task on VGGFace2 and BioSecure databases. In a recent study \cite{RecOrFQ19}, a deep CNN is utilized to calculate facial quality metric, and it is trained by using faces augmented with brightness, contrast, blur, occlusion, pose change and face matching score obtained from the distance between input image and reference image via PLDA. This work introduces use of occlusion during FQA training on the other hand just the eyes-mouth-nose landmark regions are covered in an unrealistic way that do not fit with surveillance scenarios. 

\begin{figure*}[t]
\setlength\tabcolsep{1pt}%
\begin{tabularx}{\textwidth}{@{}c*{4}{C}@{}}
 &
   \includegraphics[ width=\linewidth, height=\linewidth, keepaspectratio]{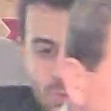} &
   \includegraphics[ width=\linewidth, height=\linewidth, keepaspectratio]{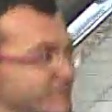}
   &
   \includegraphics[ width=\linewidth, height=\linewidth, keepaspectratio]{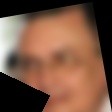} &
   \includegraphics[ width=\linewidth, height=\linewidth, keepaspectratio]{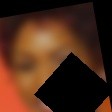}
   \\
   & Blur\cite{Lienhard-2015}: 0.51 & Blur\cite{Lienhard-2015}: 0.40 & 
   Blur\cite{Lienhard-2015}:0.60 & Blur\cite{Lienhard-2015}: 0.57
   \\
   & Contrast\cite{Lienhard-2015}: 0.42 & Contrast\cite{Lienhard-2015}: 0.67 & Contrast\cite{Lienhard-2015}: 1.0 & Contrast\cite{Lienhard-2015}: 1.0
   \\
   & Proposed: 0.19 & Proposed: 0.11 & Proposed: 0.03 & Proposed: 0.04
   \\
   & FR Score: 0.38 & FR Score: 0.15 & FR Score: 0.14 & FR Score: -0.01
   \\
\end{tabularx}
\caption{The face quality assessment should be highly correlated with the face recognition scores for surveillance scenarios. } \label{fig:occlusion}
\end{figure*}

\section{Motivation}
State-of-the-art has focused on developing algorithms that tackle FQA without much concern on computational analysis that results in complex networks with high execution times. On the other hand, FQA is a tool that is mostly utilized on edge devices incorporated with face detection, tracking and face image post-processing. Therefore, it should consume moderate computational load that is almost seamless to achieve real-time performance on the cameras.

In a complete face recognition system, the main target is to achieve high face matching accuracy with low false alarm rates. In that manner, FQA should associate with face recognition in such a way that high face matching scores are accompanied with high face quality measures. This helps to eliminate low quality faces and matching is only performed on limited number of faces that are more likely to be matched.

Classical image aesthetics measurements can easily be deployed on the edge devices for FQA however they do not provide sufficient accuracy in terms of face recognition especially for the surveillance scenarios including severe occlusions and low local contrast, as in Figure \ref{fig:occlusion}. Focusing on the face landmark points and calculating image quality on those regions improves FQA for motion blur and low contrast faces but still lacks occlusion cases which are very common in surveillance.

Most of the FQA methods do not consider face distortion scenarios observed in surveillance during the augmentation steps of model training. In that manner, artificial blur and occlusion are important tools for simulating scenarios that can be faced in surveillance applications as shown in Figure \ref{fig:occlusion}. Recently, \cite{Occ_aware_FD} has introduced occlusion as a tool during augmentation of their training data, while their aim is to extract occluding regions such as mask or sun-glasses in a face image at the detection step rather than FQA.

In this study, we tackle FQA by proposing a technique following a recognition guided motivation. We utilize an extra output layer in a well known landmark extraction network without additional complexity. Besides, we exploit random rotation, Gaussian blur and occlusion as augmentation steps in the training phase, that helps to model the distortion scenarios in realistic surveillance FR. Finally, through extensive experiments, we illustrate that a simple layer extension is quite sufficient to measure face quality in surveillance scenarios that has correlation with recognition as illustrated in Figure \ref{fig:occlusion}.

\section{Recognition Guided Face Quality Assessment}
\subsection{FR system overlook}
\begin{figure}[t]
  \includegraphics[width=\linewidth]{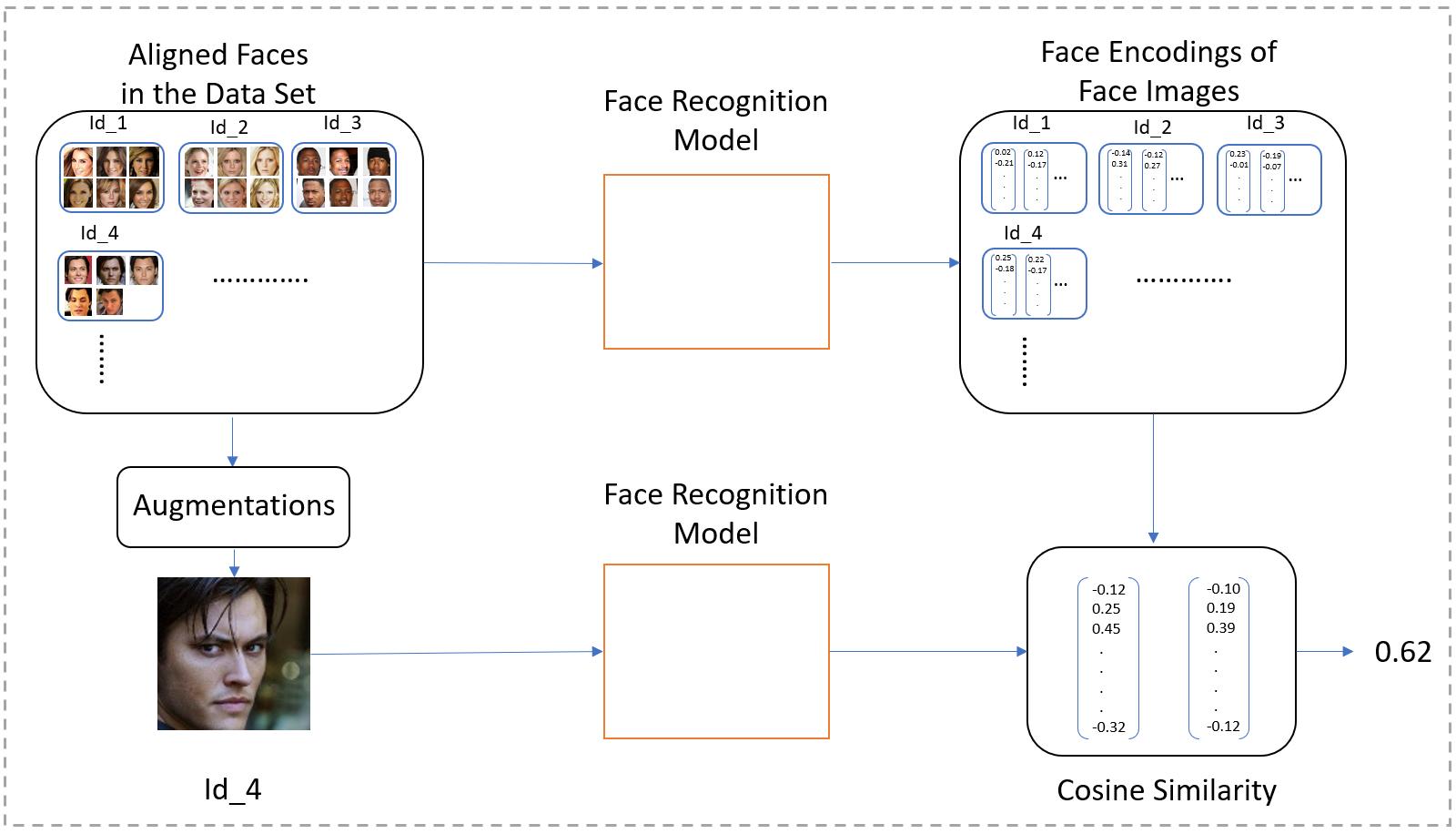}
  \caption{Face recognition system used for extracting face similarity scores} 
  \label{fig:fr_model}
\end{figure}

FQA is an intermediate step along with face detection, tracking and alignment which is embedded in the landmark extraction network as an extra output for the sake of computational efficiency. The landmark extraction network's weights are frozen during the training stage and the extra added layer is trained by face recognition similarity scores that are considered as face quality score. In order to obtain these scores, we employed ArcFace\cite{ArcFace2018} that is one of the state-of-the-art face recognition methods. It is also important to note that the proposed approach does not rely on ArcFace\cite{ArcFace2018} and the recognition model can be replaced by another one as will be illustrated in the experimental results. As a CNN architecture, ResNet50\cite{ResNet2016} is wielded with output embedding size of 512. Next, CelebA dataset\cite{CelebA2015} is utilized to train FQA layer along with cosine similarity metrics applied on the Arcface embeddings. The overall face recognition system that is exploited in order to obtain face similarity scores is shown in Figure \ref{fig:fr_model}. 

\subsection{Network Structure}

\begin{figure}[t]
  \includegraphics[width=\linewidth]{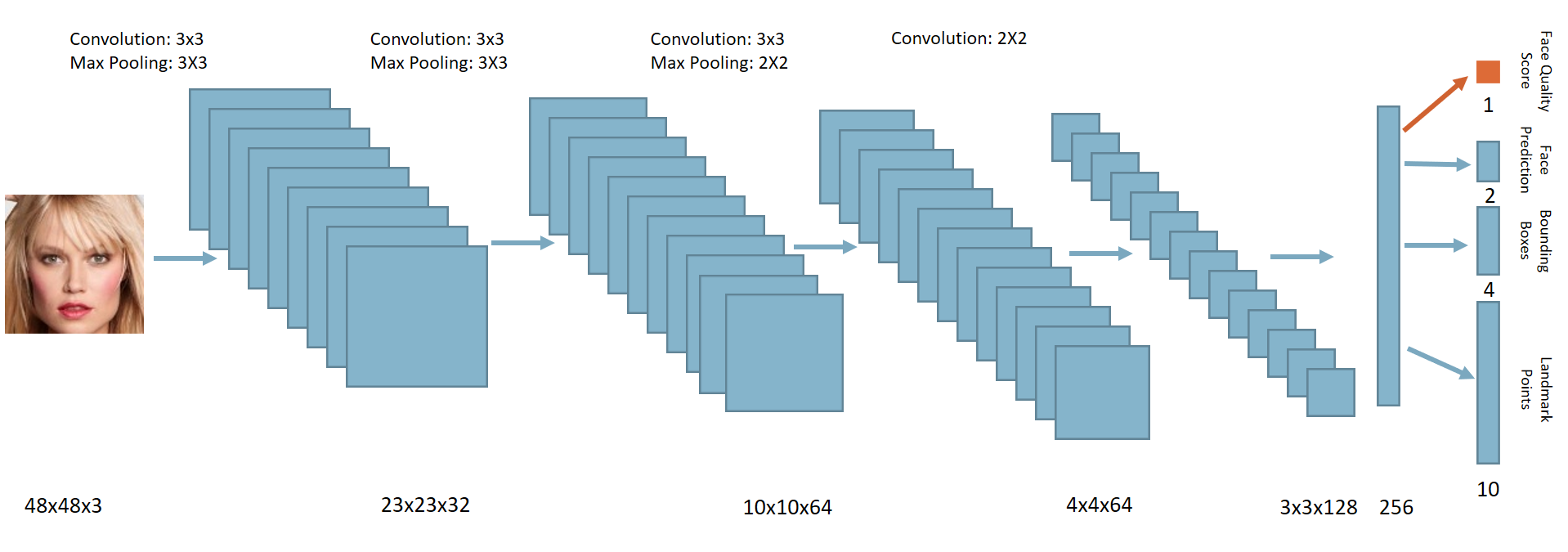}
  \caption{Modified O-Net architecture}
  \label{fig:monet}
\end{figure}

Multi-task Cascaded Convolutional Neural Network(MTCNN) \cite{Mtcnn2016}, is one of the well known face detection and facial landmark points extraction networks. It consists of 3 cascaded stages Proposal Network(P-Net), Refine Network(R-Net), Output Network(O-Net) in order to boost both performance of face detection and face alignment. P-Net produces proposal bounding box windows and performs bounding box regression, R-net rejects false proposals and refine bounding boxes, and O-net also further refines bounding boxes and produces 5 facial landmark coordinates. In designing FQA network, our first aim is to avoid increase in computational load; hence we add only one extra node to the output layer of O-Net instead of creating a new CNN. In addition, we remove the fully connected layer which is used for predicting face class since we do not use the face detector output of this model for deployment and we input only face images to the modified o-net, aka \textit{monet}. Adding an extra node into the overall network architecture as shown in Figure \ref{fig:monet} enables the estimation of face quality score with a negligible computational cost. 

\subsection{Augmentation}

There are different types of distortions in surveillance scenarios affecting quality of captured faces as well as face recognition performance. In that manner, change in head orientation, motion blur and occlusion are the most common cases in such scenarios where the subjects are not aware of face recognition and behave naturally. Thus, we apply random rotation, random Gaussian blur and random occlusion to face images in order to duplicate these realistic distortions. 

\subsubsection{Rotation:}

In real-life scenes, head orientation can change naturally among consecutive frames. Even though faces are aligned to compensate rolling of the faces and enforced to have horizontally aligned eye locations, this process depends on detection of landmarks' points, which might cause some errors. In order to take its effect into account, we rotate faces between -15 and 15 degrees randomly. 

\subsubsection{Blurring:}

Surveillance cameras are generally zoomed for capturing faces with more detail (ideally eyes are separated by 60 pixels). Besides, people move naturally without any attempt to make their faces recognizable in front of a device. These two facts result in blur which definitely drops face recognition performance. Thus, we add some random Gaussian blur to the cropped faces of 112x112 resolution with changing kernel sizes between 3 and 21.

\subsubsection{Occlusion:}
Occlusion in faces is another cause that decreases face recognition performance in surveillance scenes. In the augmentation, we add random occlusions to face images by creating a rectangle with a maximum size of one-fourth of the face image size and roll it randomly. Then, the rectangle is placed to a random location in the image, which will occupy at most 25 percent of the image. 

\begin{figure*}[t]
\setlength\tabcolsep{2pt}%
\begin{tabularx}{\textwidth}{@{}c*{8}{C}@{}}
 & Original & Blur &  Rotation & Occlusion &  BRO & BRO & BRO & BRO \\
 &
   \includegraphics[ width=\linewidth, height=\linewidth, keepaspectratio]{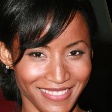} &
   \includegraphics[ width=\linewidth, height=\linewidth, keepaspectratio]{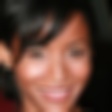} &
   \includegraphics[ width=\linewidth, height=\linewidth, keepaspectratio]{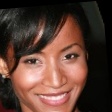} &
   \includegraphics[ width=\linewidth, height=\linewidth, keepaspectratio]{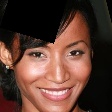} &
   \includegraphics[ width=\linewidth, height=\linewidth, keepaspectratio]{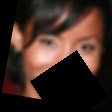} &
   \includegraphics[ width=\linewidth, height=\linewidth, keepaspectratio]{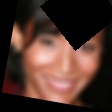} &
   \includegraphics[ width=\linewidth, height=\linewidth, keepaspectratio]{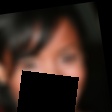} &   
   \includegraphics[ width=\linewidth, height=\linewidth, keepaspectratio]{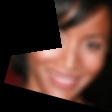} \\
 &
   \includegraphics[ width=\linewidth, height=\linewidth, keepaspectratio]{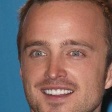} &
   \includegraphics[ width=\linewidth, height=\linewidth, keepaspectratio]{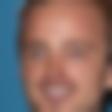} &
   \includegraphics[ width=\linewidth, height=\linewidth, keepaspectratio]{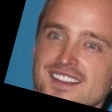} &
   \includegraphics[ width=\linewidth, height=\linewidth, keepaspectratio]{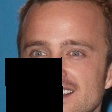} &
   \includegraphics[ width=\linewidth, height=\linewidth, keepaspectratio]{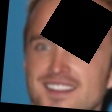} &
   \includegraphics[ width=\linewidth, height=\linewidth, keepaspectratio]{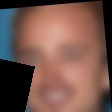} &
   \includegraphics[ width=\linewidth, height=\linewidth, keepaspectratio]{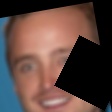} &   
   \includegraphics[ width=\linewidth, height=\linewidth, keepaspectratio]{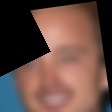} \\
 &
   \includegraphics[ width=\linewidth, height=\linewidth, keepaspectratio]{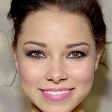} &
   \includegraphics[ width=\linewidth, height=\linewidth, keepaspectratio]{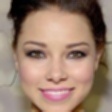} &
   \includegraphics[ width=\linewidth, height=\linewidth, keepaspectratio]{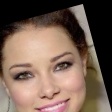} &
   \includegraphics[ width=\linewidth, height=\linewidth, keepaspectratio]{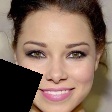} &
   \includegraphics[ width=\linewidth, height=\linewidth, keepaspectratio]{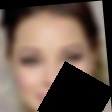} &
   \includegraphics[ width=\linewidth, height=\linewidth, keepaspectratio]{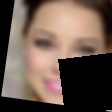} &
   \includegraphics[ width=\linewidth, height=\linewidth, keepaspectratio]{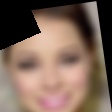} &   
   \includegraphics[ width=\linewidth, height=\linewidth, keepaspectratio]{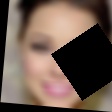} \\
\end{tabularx}
\caption{Blur, rotation and occlusion augmentations are utilized in the training to simulate real life surveillance data, images are taken from CelebA dataset \cite{CelebA2015}} \label{fig:augmentation}
\end{figure*}

Some examples of augmentations on CelebA dataset are shown in \ref{fig:augmentation} for three different faces. In the first three columns, single attacks of blurring, rotation and occlusion are given, while the rest four columns correspond to random combinations of these three.

\subsection{Training Dataset}

\subsubsection{CelebA Faces:}  

Large-scale CelebFaces Attributes(CelebA) dataset\cite{CelebA2015} is one of the most suitable data sets for training FQA networks due to several reasons. First of all, it has identity labels in order to extract face similarity scores required for training our FQA network. Secondly, it contains more than 200k face images with different resolutions, pose variations and background. Finally, it has also face bounding boxes, 5 facial landmark points and many attribute labels (around 40). \cite{CelebA2015} consists of 202,599 face images from 10,177 different identities and we have used training part of CelebA which is composed of 162770 face images for training our FQA layer.

\subsection{Network Training}

\textbf{Fig. 8.} MTCNN  Pipeline The modified version of O-Net architecture is trained on CelebA dataset with various alternatives of augmentation methods as rotation, blur, occlusion and rotation+blur+occlusion. This enables the analyses of the effect of augmentations and their representation capability. Training batch size is set to 128 and training started with learning rate of 0.01. CNN training is performed on a workstation with Nvidia GTX 1080Ti GPU by using MXNet framework \cite{MXNet2015}. 

\section{Experimental Results}

The experiments are conducted in three main sessions, where in the first section comprehensive tests are performed for the proposed approach with various alternatives on the training data. In the second section, proposed approach is compared with state of the art face quality assessment as well as classical image quality metrics. In the final section we run the proposed approach within a complete pipeline including face detection tracking and alignment on a well known mobile GPUs (Nvidia Jetson TX2). Before getting into the details of the experiments, we would like to introduce the datasets utilized throughout the experiments.

\subsection{Dataset Analyses}

\subsubsection{LFW Dataset:}

Labeled Faces in the Wild(LFW) \cite{LFW2007},\cite{LFW2014} is a well-known benchmark dataset for face verification. It contains 13,233 images from 5749 people and they are suitable for research on unconstrained face recognition. In this work, we utilize LFW for comparing average face quality score of original pairs with those of Gaussian blurred and/or occluded faces. The face pairs are taken from \cite{ArcFace2018} Github repository, which has 3,000 true matched pairs and 3,000 false matched pairs. In order to compare the change in average face quality scores with distortions, we utilize the true matched pairs.

\subsubsection{Color FERET Database:}

The Face Recognition Technology (FERET) \cite{FERET1998},\cite{FERET2000} database consists of 14,126 images from 1199 individuals. The database has been collected by different organizations in highly controlled environment. It has been created for developing face recognition technology via new algorithms and techniques. We take advantage of FERET database for evaluating our FQA method as in LFW. For each image in the database, a match pair is assigned through face matching via \textit{Arcface} and \textit{facenet} under different scenarios.

\subsubsection{Georgia Tech Face Database:}

Georgia Tech Face Database\cite{GeorgiaTech} is comprised of 750 images taken from 50 individuals. Face images have been collected under different illumination conditions with different facial expressions and orientations. The best matches are assigned for each image which could alter according to the simulated attacks during experiments.

\subsubsection{Realistic Surveillance Data:}

Apart from the previously described three datasets, we also exploit a realistic surveillance scenario where 15 different people walk in a crowd in various scenes. We have recorded videos on three different sites, \textit{lab}, \textit{indoor\_low} and \textit{indoor\_high}. The last two differentiate with respect to the location of camera, low indicates camera location at 2 meters while high indicates camera located in 4 meters with high zoom level. This data provides observation of the FQA performance under real scenarios where natural motion blur and occlusion as well as different head orientations and natural cases are observed including yawning, phone call, drinking, looking down. 

\subsection{Algorithm Tests}

The proposed faceness measure approach has been tested on four different sets as referenced recently. We apply artificial attacks on the original data (the first three) in order to simulate the real surveillance scenarios such as blur and occlusion that are very common and measure the face matching scores via \textit{Arcface} \cite{ArcFace2018}. It is important to note that, we do not apply any attacks on the realistic surveillance data which has natural distortions. In that manner, we analyze three scenarios with attacks including only Gaussian blur, only partial occlusion and both. Gaussian blur is applied with random standard deviation, partial occlusion is applied by choosing four random points within the image and painting the covered area. Before getting into the results of the proposed method, we would like to share the average similarity values as well as the effects of simulated attacks on the \textit{Georgia}, \textit{LFW} and \textit{Feret} datasets in Table \ref{tab:dataAugEffect}. It is clear that, the face matching similarity is dropped severely (to $\textfrac{1}{3}$ of the average similarity in case of no attack) when both blur and occlusion attacks are applied to the test images. On the other hand, adding occlusion has less impact in \textit{Georgia} and \textit{LFW} compared to blurring that is vice-versa for \textit{Feret} dataset. This behaviour is valid for both face similarity metrics.

\begin{table*}[t]
\centering
\begin{center}
    \caption{The effect of attacks on different average face matching scores for different datasets}
        \begin{tabular}{c c c c c c}
            {DataSet} & Similarity & {No Attack} & {Blur} & {Occ} & Blur+Occ \\
            \hline
            \multirow{2}{*}{Georgia} & Arcface & 0.85 & 0.61 & 0.65 & 0.24 \\
            & FaceNet & 0.48 & 0.69 & 0.78 & 1.12 \\
            \hline
            \multirow{2}{*}{LFW} & Arcface & 0.71 & 0.34 & 0.53 & 0.21 \\
            & FaceNet & 0.33 & 0.62 & 0.53 & 0.74  \\
            \hline
            \multirow{2}{*}{Feret} & Arcface & 0.59 & 0.44 & 0.39 & 0.19 \\
            & FaceNet & 0.68 & 0.99 & 0.92 & 1.19 \\
            \hline
        \end{tabular}
        \label{tab:dataAugEffect}
\end{center}
\centering
\end{table*}

The performance of FQA is measured through the Pearson correlation coefficient that calculates the linearity between the measurements. In that manner, after each attack, the estimated face quality and the actual matching score are measured that are subject to the Pearson correlation. This correlation approaches to "1" as the measurements are linearly related to each other. 

We come up with four FQA models that are trained under various augmentation scenarios. \textit{Blur} model is trained on the data distorted by only Gaussian blur, \textit{Rot} model is trained by data distorted through only rotation, \textit{Occ} model is trained by only occlusion augmentation and the final model \textit{BRO} is trained by all augmentations. These four models help us to understand if the models are trained properly and behave as expected. 

These four models estimate the faceness quality of each image under three different distortions (Gaussian blur, Occlusion and both). Then for each distorted image two scores are calculated via \textit{Arcface}, the best matching among the other images in the database and self similarity with the un-distorted original version of the image. The Pearson coefficients are calculated between the matching scores and the estimated faceness quality measures. The results are given for GeorgiaTech, LFW and Feret datasets in Table \ref{tab:own_Gerogia} correspondingly. The correlation coefficients show same tendency for each dataset, where each attack is best represented by the models that are trained over the same augmentation set. The unified \textit{BRO} model has the best correlation with real matching scores for the attacks that include both blur and occlusion. Besides, it always has the second best Pearson correlations for the other attacks that are very close to the best correlations in each row. 

Among these various alternatives on attacks and datasets, \%78 of the achieved best Pearson correlations are above 0.75 which indicates very good to excellent correlation between variables. This means that the proposed framework enables high quality FQA as long as proper augmentations are applied in the training. 

For the sake of completeness we also exploited \textit{Facenet} \cite{facenet} to train and evaluate the FQA models apart from \textit{Arcface}. We calculated the Pearson correlations for two BRO models trained according to face recognition scores of different methods, namely \textit{Arcface} and \textit{FaceNet}. We utilized two face similarity measures and cross-checked to observe if the proposed approach is sensitive to the similarity metric choices in training phase. Since two metrics yield opposite behaviors for the best matches, where \textit{Arcface} tends to approach \textit{1.0} and \textit{FaceNet} tends to approach \textit{0.0} for a perfect match, the Pearson correlations are negated in the cross scenarios for the sake of simplicity. The results are given in Table \ref{tab:SimilarityCrossCheck}, where \textit{Arcface} based model training achieves superior correlation compared to \textit{FaceNet}. On the other hand, the difference between two models under the same similarity during test phase (two measures on the same row) does not exceed \%2 indicating that the proposed approach is almost independent of the face similarity metrics exploited during the train phase. It is also clear that, \textit{Arcface} provides higher Pearson correlation when it is applied in matching during tests as the results in the same columns are checked.       
\begin{table*}[t]
\centering

    \caption{Face quality prediction correlation with actual scores on GeorgiaTech, LWF and Feret face datasets}
        \begin{tabular}{c c | c c c c | c c c c | c c c c}
            \toprule
            \midrule
                \multicolumn{2}{c}{} & \multicolumn{4}{c}{Georgia} & \multicolumn{4}{c}{LFW} & \multicolumn{4}{c}{Feret}\\
                \hline
                {Attack} & Similarity & {Blur} & {Rot} & {Occ} & {BRO} & {Blur} & {Rot} & {Occ} & {BRO} & {Blur} & {Rot} & {Occ} & {BRO} \\ 
                \hline
                \multirow{2}{*}{Blur} & Match & \textbf{0.84} & 0.67 & 0.56 & \textbf{0.84} & \textbf{0.88} & 0.73 & 0.79 & 0.87 & \textbf{0.71} & 0.46 & 0.64 & 0.69\\
               & Self & \textbf{0.86} & 0.65 & 0.53 & 0.85 & \textbf{0.62} & 0.29 & 0.47 & 0.57 & \textbf{0.62} & 0.29 & 0.47 & 0.57\\
                \hline
               \multirow{2}{*}{Occlusion} & Match & 0.73 & 0.80 & \textbf{0.82} & \textbf{0.82} & 0.67 & 0.71 & \textbf{0.74} & 0.73 & 0.62 & 0.62 & \textbf{0.78} & 0.72 \\
              & Self & 0.68 & 0.76 & \textbf{0.79} & 0.78 & 0.56 & 0.55 & \textbf{0.75} & 0.66 & 0.56 & 0.55 & \textbf{0.75} & 0.66   \\
                \hline
                \multirow{2}{*}{Blur+Occ} & Match & 0.71 & 0.73 & 0.68 & \textbf{0.79} & 0.79 & 0.64 & 0.66 & \textbf{0.82} & 0.58 & 0.52 & 0.64 & \textbf{0.69}  \\
                & Self & 0.67 & 0.70 & 0.66 & \textbf{0.77} & 0.83 & 0.67 & 0.69 & \textbf{0.86} & 0.56 & 0.48 & 0.62 & \textbf{0.66}\\
            \midrule
            \bottomrule
        \end{tabular}
    \label{tab:own_Gerogia}
\centering
\end{table*}

\begin{table}[h!]
  \begin{center}
    \caption{Pearson correlation scores of face quality scores from BRO models trained by different FR models with respect to face similarity scores obtained from different FR models}
    \begin{tabular}{c c | c c}
            \toprule
            \midrule
                \multicolumn{2}{c}{Pearson Corr.} & \multicolumn{2}{c}{BRO Models}\\
                \hline
                {Dataset} & Face Recognition Models & {Arcface} & {FaceNet} \\ 
                \hline
                \multirow{2}{*}{GeorgiaTech} & Arcface &  \textbf{0.787} & 0.768 \\
               & FaceNet & 0.732 & 0.730 \\
                \hline
               \multirow{2}{*}{LFW} & Arcface & \textbf{0.819} & 0.807 \\
               & FaceNet & 0.759 & 0.749 \\
                \hline
                \multirow{2}{*}{Feret} & Arcface & \textbf{0.686} & 0.676 \\
               & FaceNet & 0.609 & 0.583 \\
            \midrule
            \bottomrule
    \end{tabular}
    \label{tab:SimilarityCrossCheck}
  \end{center}
\end{table}

\subsection{Comparison with state-of-the-art}
In this section, we compare the proposed FQA method trained under \textit{BRO} augmentation with a recent technique \textit{FaceQnet} \cite{HernandezOrtega2019} developed for FQA as well as two classical measures blur and contrast proposed in \cite{Lienhard-2015}. We utilize both face similarity metrics and two BRO models trained through these metrics. As in the previous section, we measure correlation of the actual similarity and the estimated FQA by the Pearson correlation.   

\begin{table*}[!h]
\centering
    \caption{Pearson correlation of proposed, FaceQnet, blur and occlusion metrics with respect to ArcFace/FaceNet matching scores }
        \begin{tabular}{c c | c c c c}
            \toprule
            \midrule
                {Attack} & {Dataset} & {FQA} & {FaceQnet} & {Blur} & {Contrast}\\
                \hline
                \multirow{3}{*}{Blur} & Georgia & \textbf{0.84}/\textbf{0.73} & 0.78/0.52 & 0.58/0.59 & 0.21/0.29\\
               & LFW & \textbf{0.87}/\textbf{0.80} & 0.69/0.69 & 0.79/0.68 & 0.22/0.19\\
               & Feret & 0.69/0.65 & \textbf{0.71}/\textbf{0.73} & 0.42/0.37 & 0/0.08\\
                \hline
               \multirow{3}{*}{Occlusion} & Georgia & \textbf{0.82}/0.73 & 0.78/\textbf{0.74} & 0.03/0.02 & 0.07/0.06\\
              & LFW & 0.72/0.68 & \ \textbf{0.73}/\textbf{0.72} & 0.03/0.02 & 0/0\\
              & Feret & \textbf{0.72}/0.71 &  0.71/\textbf{0.79} & 0.06/0.02 & 0.01/0.12\\
                \hline
                \multirow{3}{*}{Blur+Occ} & Georgia & \textbf{0.79}/\textbf{0.73} & 0.71/0.70 & 0.57/0.55 & 0.05/0\\
                & LFW & \textbf{0.82}/\textbf{0.75} & 0.62/0.65 & 0.79/0.69 & 0.06/0.04\\
                & Feret & \textbf{0.69}/0.58 & 0.66/\textbf{0.70} & 0.49/0.03 & 0.03/0.06\\
            \midrule
            \bottomrule
        \end{tabular}
    \label{tab:sota_comparsion}
\centering
\end{table*}

\begin{table*}[h]
\centering
    \caption{Pearson correlations of FQA methods with respect to FR models on surveillance videos}
        \begin{tabular}{c c | c c c c}
            \toprule
            \midrule
                \multicolumn{2}{c}{} & \multicolumn{4}{c}{FQA Models}\\
                \hline
                {Dataset} & Similarity & {BRO} & {FaceQnet} & {Blur} & {Contrast} \\ 
                \hline
                \multirow{2}{*}{Lab} & Arcface & \textbf{0.76} & 0.73 & 0.22 & 0.05  \\
               & FaceNet & 0.49 & \textbf{0.62} & 0.22 & 0.07 \\
                \hline
               \multirow{2}{*}{Indoor\_low} & Arcface & \textbf{0.79} & 0.77 & 0.22 & 0.12 \\
              & FaceNet & 0.64 & \textbf{0.67} & 0.24 & 0.27  \\
                \hline
                \multirow{2}{*}{Indoor\_high} & Arcface & \textbf{0.59} & 0.54 & 0.11 & 0.04 \\
                & FaceNet & \textbf{0.69} & 0.65 & 0.08 & 0.06 \\
            \midrule
            \bottomrule
        \end{tabular}
    \label{tab:surveillance}
\centering
\end{table*}

The Pearson correlations are given in Table \ref{tab:sota_comparsion} for each FQA method under three distortion attacks on three different data sets calculated for two different face matching scores. It is obvious that proposed approach provides the highest correlation with the actual face matching scores for 11 cases out of 18, where the rest is provided by \textit{FaceQnet}. The classical measures fail to predict the FQA in terms of face matching. However, under the Gaussian blur and joint attack scenario the \textit{blur} metric has significant correlation due to severe decrease in sharpness of the face images. \textit{Contrast} can not provide any correlation with the face recognition score especially under the occlusion attacks. These observation validate the motivation of this study that classical FQA are not sufficient to model face matching similarity. 

\begin{figure*}[t]
\setlength\tabcolsep{2pt}%
\begin{tabularx}{\textwidth}{@{}c*{8}{C}@{}}
 &
   \includegraphics[ width=\linewidth, height=\linewidth, keepaspectratio]{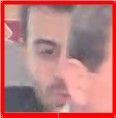} &
   \includegraphics[ width=\linewidth, height=\linewidth, keepaspectratio]{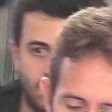}
   &
    \includegraphics[ width=\linewidth, height=\linewidth, keepaspectratio]{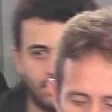}
   &
   \includegraphics[ width=\linewidth, height=\linewidth, keepaspectratio]{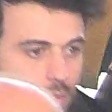}
   &
   \includegraphics[ width=\linewidth, height=\linewidth, keepaspectratio]{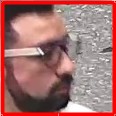}
   &
   \includegraphics[ width=\linewidth, height=\linewidth, keepaspectratio]{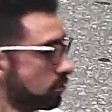}
   &
   \includegraphics[ width=\linewidth, height=\linewidth, keepaspectratio]{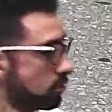}
   &
   \includegraphics[ width=\linewidth, height=\linewidth, keepaspectratio]{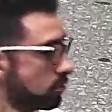} \\
   &
   \includegraphics[ width=\linewidth, height=\linewidth, keepaspectratio]{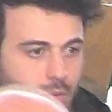} &
   \includegraphics[ width=\linewidth, height=\linewidth, keepaspectratio]{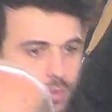}
   &
    \includegraphics[ width=\linewidth, height=\linewidth, keepaspectratio]{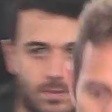}
   &
   \includegraphics[ width=\linewidth, height=\linewidth, keepaspectratio]{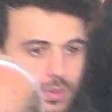}
   &
   \includegraphics[ width=\linewidth, height=\linewidth, keepaspectratio]{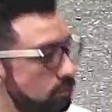}
   &
   \includegraphics[ width=\linewidth, height=\linewidth, keepaspectratio]{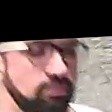}
   &
   \includegraphics[ width=\linewidth, height=\linewidth, keepaspectratio]{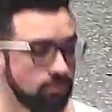}
   &
   \includegraphics[ width=\linewidth, height=\linewidth, keepaspectratio]{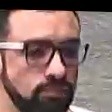}
   \\
   &
   \includegraphics[ width=\linewidth, height=\linewidth, keepaspectratio]{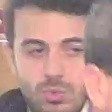} &
   \includegraphics[ width=\linewidth, height=\linewidth, keepaspectratio]{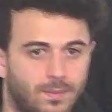}
   &
    \includegraphics[ width=\linewidth, height=\linewidth, keepaspectratio]{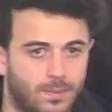}
   &
   \includegraphics[ width=\linewidth, height=\linewidth, keepaspectratio]{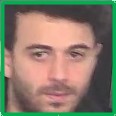}
   &
   \includegraphics[ width=\linewidth, height=\linewidth, keepaspectratio]{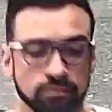}
   &
   \includegraphics[ width=\linewidth, height=\linewidth, keepaspectratio]{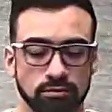}
   &
   \includegraphics[ width=\linewidth, height=\linewidth, keepaspectratio]{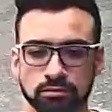}
   &
   \includegraphics[ width=\linewidth, height=\linewidth, keepaspectratio]{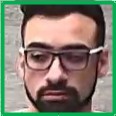} \\
   & \textcolor{red}{FaceQnet:0.44} & & \textcolor{mygreen}{FaceQnet:0.62} & & \textcolor{red}{FaceQnet:0.49} & &\textcolor{mygreen}{FaceQnet:0.60} & \\
   &\textcolor{red}{Blur:0.51} & & \textcolor{mygreen}{Blur:0.53} & & \textcolor{red}{Blur:0.36} & &  \textcolor{mygreen}{Blur:0.37} &
   \\
   & \textcolor{red}{Contrast:0.42} & & \textcolor{mygreen}{Contrast:0.58} & & \textcolor{red}{Contrast:0.75} & & \textcolor{mygreen}{Contrast:0.82} &
   \\
   & \textcolor{red}{Proposed:0.19} & & \textcolor{mygreen}{Proposed:0.63} & & \textcolor{red}{Proposed:0.20} & & \textcolor{mygreen}{Proposed:0.68} &
   \\
   & \textcolor{red}{Similarity:0.40} & & 
   \textcolor{mygreen}{Similarity:0.70} & & 
   \textcolor{red}{Similarity:0.39} & & 
   \textcolor{mygreen}{Similarity:0.63} &
   \\
\end{tabularx}
\caption{The sorted faces of two individuals from our surveillance videos, red indicates the lowest quality face while green indicates the highest quality face of the tracks. The estimated face quality measures for the best and worst face crops are given per FQA model. } \label{fig:tf}
\end{figure*}

\begin{figure*}[t]
\setlength\tabcolsep{2pt}%
\begin{tabularx}{\textwidth}{@{}c*{7}{C}@{}}
 &
   \includegraphics[ width=\linewidth, height=\linewidth, keepaspectratio]{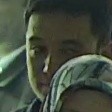} &
   \includegraphics[ width=\linewidth, height=\linewidth, keepaspectratio]{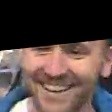}
   &
    \includegraphics[ width=\linewidth, height=\linewidth, keepaspectratio]{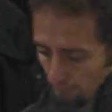}
   &
   \includegraphics[ width=\linewidth, height=\linewidth, keepaspectratio]{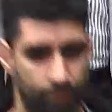}
   &
   \includegraphics[ width=\linewidth, height=\linewidth, keepaspectratio]{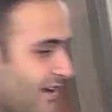}
   &
   \includegraphics[ width=\linewidth, height=\linewidth, keepaspectratio]{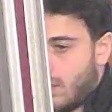}
   &
   \includegraphics[ width=\linewidth, height=\linewidth, keepaspectratio]{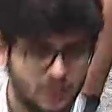} \\
   
   & 0.35/0.32 & 0.38/0.35 & 0.42/0.31 & 0.32/0.44 & 0.55/0.41 & 0.45/-0.01 & 0.40/0.44 \\
   &
   \includegraphics[ width=\linewidth, height=\linewidth, keepaspectratio]{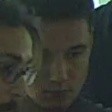} &
   \includegraphics[ width=\linewidth, height=\linewidth, keepaspectratio]{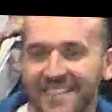}
   &
    \includegraphics[ width=\linewidth, height=\linewidth, keepaspectratio]{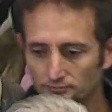}
   &
   \includegraphics[ width=\linewidth, height=\linewidth, keepaspectratio]{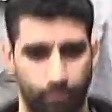}
   &
   \includegraphics[ width=\linewidth, height=\linewidth, keepaspectratio]{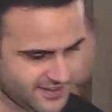}
   &
   \includegraphics[ width=\linewidth, height=\linewidth, keepaspectratio]{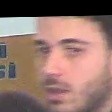}
   &
   \includegraphics[ width=\linewidth, height=\linewidth, keepaspectratio]{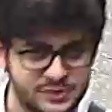} \\
   &
   0.41/0.40 & 0.63/0.54 & 0.57/0.68 & 0.52/0.54 & 0.59/0.58 & 0.54/0.36 & 0.56/0.61 \\
   &
   \includegraphics[ width=\linewidth, height=\linewidth, keepaspectratio]{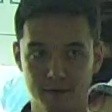} &
   \includegraphics[ width=\linewidth, height=\linewidth, keepaspectratio]{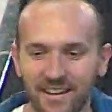}
   &
    \includegraphics[ width=\linewidth, height=\linewidth, keepaspectratio]{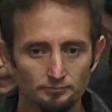}
   &
   \includegraphics[ width=\linewidth, height=\linewidth, keepaspectratio]{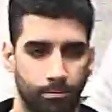}
   &
   \includegraphics[ width=\linewidth, height=\linewidth, keepaspectratio]{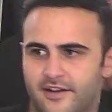}
   &
   \includegraphics[ width=\linewidth, height=\linewidth, keepaspectratio]{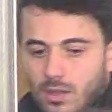}
   &
   \includegraphics[ width=\linewidth, height=\linewidth, keepaspectratio]{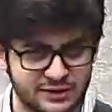}
   \\
   & 0.62/0.52 & 0.71/0.69 & 0.66/0.78 & 0.66/0.57 & 0.71/0.78 & 0.66/0.68 & 0.65/0.66 \\
\end{tabularx}
\caption{The FQA of the proposed method and the face similarity (\textit{Arcface}) values are given for ordered faces of seven individuals.} \label{fig:3sets}
\end{figure*}

We extend this comparison for the realistic data that we have collected as mentioned in section 5.1. In this scenario, the face images are gathered from surveillance videos to observe the real cases including occlusion and motion blur without any artificial attacks. In Table \ref{tab:surveillance}, the Pearson correlation of each are given for three different sites. It is clear that proposed approach has competitive results with \textit{FaceQnet}, where it has higher correlation when \textit{Arcface} is utilized for face similarity. The correlations for both methods are very close to each other. The sorted face crops of two individuals are given in Figure \ref{fig:tf} that are gathered by consecutive face detection and tracking that yield same track ID for each person. The FQ estimations of the worst (red rectangle) and the best (green rectangle) faces are given below the Figure \ref{fig:tf}, where proposed approach has the best discrimination between the best and the worst images compared to the other techniques. Face quality estimates are significantly increased by the proposed method when faces are observed crisp and clear, while these values drop severely under motion blur and occlusion. Further results are given in Figure \ref{fig:3sets} for surveillance scenario, where the estimated face quality and the actual recognition scores are given for each crop. It is obvious that the proposed approach is consistent with the actual face similarity metrics and provides valid prejudgement of faces before recognition.

\subsection{Edge deployment}
The proposed FQA approach is incorporated with face detection, tracking and alignment to observe if it can eliminate the low quality faces and highlight ideal faces for recognition. The implementation is conducted on well known Nvidia Jetson TX2 board as efficient, small form factor platform for edge deployment. We exploit Pelee \cite{NIPS2018_Pelee} that is a recent and efficient single shot detector for the initial detection of faces with input size of \textit{304x304}. The faces are tracked by bi-directional bounding box association, so each individual is assigned to a unique ID. The proposed FQA method is applied in conjunction with facial landmark extraction to each face in order to provide landmark points and the face quality estimate simultaneously. 
The faces are aligned for recognition and the face crops are sorted for each track ID with respect to estimated face qualities. This optimized pipeline enables real-time processing of surveillance videos. The average execution times of this application are given as \textit{15 ms}, \textit{1.0 ms}, \textit{1.8 ms} for initial face detection, bi-directional tracking and landmark extraction per face consecutively. We have also included \textit{FaceQnet} as an alternative FQA within the pipeline, where the average execution time is around \textit{19.5 ms}. It is important to note that proposed FQA technique does not require additional complexity since it is integrated into landmark extraction framework. On the other hand, \textit{FaceQnet} requires as higher computational burden as the initial face detection that introduces an overhead to the full system. Though, the proposed method is an efficient alternative for the state-of-the-art enabling consistent FQA in no time.  

\section{Conclusions}

In this paper, we propose an efficient FQA method that is incorporated with landmark extraction framework by an additional single fully connected layer. The features extracted for landmark detection are reused to regress the matching scores. In order to simulate real life scenarios of surveillance cameras, that make recognition harder, we apply rotation, blur, and occlusion attacks as data augmentation in the training of FQA layer. In depth analyses indicate that the proposed single layer extension as well as surveillance guided augmentation yield high correlation between FQA and face recognition scores. This enables quite efficient selection of best representative face images on the edge for each face track and decrease the number of face images transferred to the recognition servers. This does not only ease the network load of the face detection cameras but also potentially increases the accuracy and rate of the recognition server. Although for fixed detection and recognition networks, the introduced FQA layer provides state-of-the-art quality evaluation, further fine tuning and/or end-to-end training of detection and recognition networks with realistic augmentation attacks is the subject of future work.



\clearpage
%
%
\bibliographystyle{unsrt}
\bibliography{eccv2020submission}
\end{document}